\documentclass[lettersize,journal]{IEEEtran}
\usepackage{amsmath,amsfonts}
\usepackage[linesnumbered,boxed,ruled,commentsnumbered]{algorithm2e}
\usepackage{array}
\usepackage[caption=false,font=normalsize,labelfont=sf,textfont=sf]{subfig}
\usepackage{textcomp}
\usepackage{stfloats}
\usepackage{url}
\usepackage{verbatim}
\usepackage{graphicx}
\usepackage{cite}
\usepackage{hyperref}
\hypersetup{hidelinks,
	colorlinks=true,
	allcolors=black,
	pdfstartview=Fit,
	breaklinks=true}
\begin{document}

\title{Multi-target Robot Navigation in Crowds Based on History Interaction Information}

\author{Xinyi Yu, Jianan Hu, Yuehai Fan, Wancai Zheng, Linlin Ou
	\thanks{}
	\thanks{}}

\markboth{Journal of \LaTeX\ Class Files,~Vol.~14, No.~8, August~2021}%
{Shell \MakeLowercase{\textit{et al.}}: A Sample Article Using IEEEtran.cls for IEEE Journals}


\maketitle

\begin{abstract}
Robot navigation in dynamic environments shared with humans is an important but challenging task, which suffers from performance deterioration as the crowd grows. In this paper, multi-target robot navigation approach based on deep reinforcement learning is proposed, which can reason about more comprehensive relationships among all agents (robot and humans). Specifically, the next position point is planned for the robot by introducing history interaction information in our work. Firstly, based on subgraph network, the history information of all agents is aggregated before encoding interactions through a graph neural network, so as to improve the ability of the robot to anticipate the future trajectories of humans. Further consideration, in order to reduce the probability of unreliable next position points, the selection module is designed after policy network in the reinforcement learning framework. In addition, the next position point generated from the selection module satisfied the task requirements better than that obtained directly from the policy network. The experiments demonstrate that our approach outperforms state-of-the-art approaches in terms of both success rate and collision rate, especially in crowded human environments.
\end{abstract}


\section{Introduction}
\IEEEPARstart{W}{ith} the rapid development of robotics and artificial intelligence, the robots are required to navigate through crowded human environments, such as the guidance robots in the airports \cite{kayukawa2019bbeep}. Robot navigation in dynamic environments shared with humans still remains challenging\cite{chen2019crowd}. On the one hand, the behavior of humans is complex due to the influence of intrinsic intention, external environment and social rules \cite{rudenko2020human}. On the other hand, the explicit communication between robots and humans is often impractical. Therefore, the robots need to reason and anticipate the evolution of humans while navigating in a social etiquette way \cite{fong2003survey,kruse2013human}.

Earlier studies regard obstacles as static models \cite{fox1997dynamic} or simple dynamic models \cite{van2011reciprocal,van2011re}, which is incompatible with the complex behavior of humans. Many research efforts trajectory prediction of humans and planning based on intention inference\cite{aoude2013probabilistically,unhelkar2015human,kim2013predicting}. However, separating prediction and planning in these research will cause the freezing robot problem\cite{trautman2010unfreezing}, because the trajectory prediction will extend to the whole space, resulting in no plausible path for the robot\cite{chen2020robot}. To address these issues, a key solution is to consider the interactions among all agents when the robot moves.

Traditional methods model interactions based on hand-crafted models\cite{helbing1995social,ferrer2013robot,mehta2016autonomous,mavrogiannis2019multi}. However, the performance of these methods depends on the accuracy of the interaction models. Furthermore, the methods based on imitation learning learn the relationships between robot and humans from human data set\cite{alahi2016social,gupta2018social}. However, the performance of these methods depends on the scale and quality of the human data set. 

Recently, deep reinforcement learning (DRL) has been used successfully to learn efficient policies that implicitly encode the interactions and cooperation between robot and humans\cite{chen2017socially,chen2017decentralized,everett2018motion,chen2020relational}. In existing DRL methods, the collective impact of the crowd is usually obtained by combining pairwise interactions through a maximin operator \cite{chen2017decentralized} or LSTM \cite{everett2018motion}, which fails to fully represent all interactions. More advanced, Graph Neural Networks (GNNs) \cite{chen2020robot,chen2020relational} and pooling module based on self-attention mechanism \cite{chen2019crowd} are employed to represent fully interactions. However, there are still two problems to be  studied:
1) not only the influence of interactions at the current moment on the robot navigation, but also the influence of history information should be considered\cite{gao2020vectornet}, which boosts the ability of the robot to predict the future trajectories of humans, resulting in performance improvement in complex and crowded scenes;
2) the end-to-end methods which output velocities lack interpretation in terms of safety. Intuitively, the safer situation for the robot in the context of collision avoidance is to ensure the security of the next position point before it executes command.

\begin{figure}[!t]
	\centering
	\includegraphics[width=2.5in]{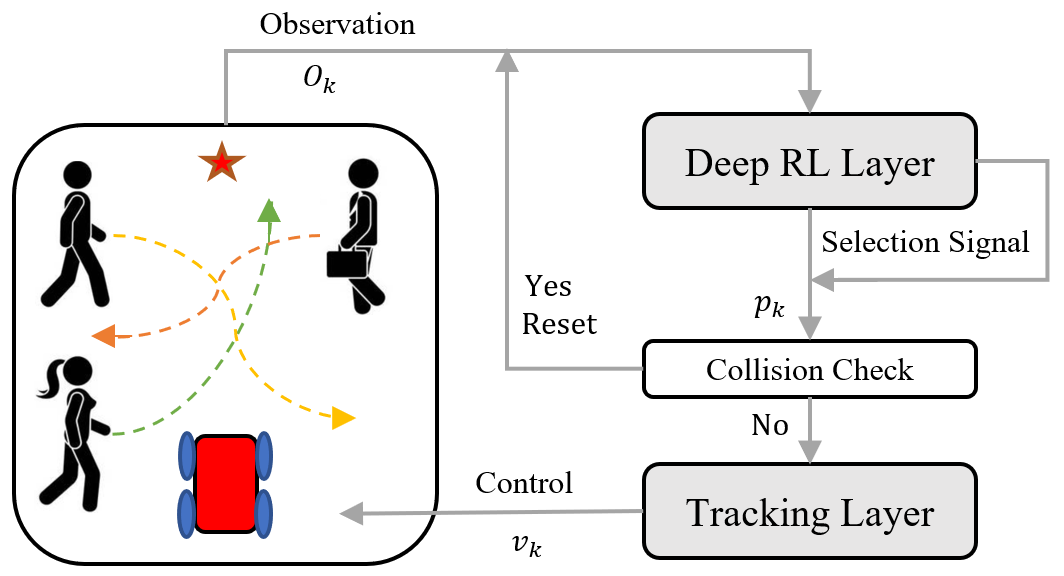}
	\caption{The hierarchical structure. The Deep RL layer will output a position point for the tracking layer to track based on the selection signal. The star represents the destination of the robot.}
	\label{fig_1}
\end{figure}
To address these issues and further improve the performance of the robot in the navigation task, we propose multi-target robot navigation approach based on DRL in this paper, which plans the next position point for the robot by considering history interaction information. In our work, we express history information through the vectorized representation and introduce a subgraph network to encode it before modeling interactions by using a GNN. Moreover, we sample multiple position points for the robot from the policy network in the reinforcement learning (RL) framework, and then develop a selection module to select the next position point adaptively, which makes the trajectory of the robot further satisfy the task requirements. In addition, the hierarchical structure in Fig. \ref{fig_1} is presented for the completeness of our method.

We evaluate our method against the state-of-the-art methods and analyze each module in our method through ablation studies in simulation environments. The experimental results demonstrate the effectiveness of each module and our method outperforms the state-of-the-art methods, especially in crowded human environments. We also demonstrate the effectiveness of our method in the real-world environments.

\section{Background}
\subsection{Related Work}
Traditional model-based robot navigation methods can be classified into reaction-based methods and trajectory-based methods \cite{everett2018motion}. Reaction-based methods use geometry or physics rules to ensure collision avoidance, such as RVO \cite{van2008reciprocal} or ORCA \cite{van2011reciprocal}. Based on definition of velocity obstacle, RVO guarantees safe and oscillation-free motions for each agent in multi-agent scenarios. Furthermore, ORCA guarantees collision-free motions for each agent without communication in a cluttered workspace. The reaction-based methods can react quickly to the current scene, but they are short-sighted due to they only look one step ahead.

Instead, trajectory-based methods \cite{aoude2013probabilistically,phillips2011sipp} make long-term trajectory prediction for agents on timescale before planning. However, the computational cost of trajectory-based methods will increase with time and prediction may lead to the freezing robot problem in the crowded environments. In addition, traditional methods model interactions to boost the social awareness of the robots, such as Social Force \cite{helbing1995social,ferrer2013robot,kretz2018some}, Interaction Gaussian Process \cite{trautman2013robot,trautman2017sparse}. The limitation of traditional methods is that they rely on hand-crafted models, which need to be adjusted in different scenarios.

Imitation learning obtains the optimal policies from human demonstration. In the context of robot navigation, the methods based on imitation leaning present an end-to-end framework to generate collision avoidance policies from raw 2D laser data \cite{long2017deep} or raw depth images \cite{tai2018socially}. The limitation of imitation leaning is that the performance of the policies learned from these methods depends on the scale and quality of human data set which is time consuming to manufacture. As in \cite{chen2019crowd}, we adopt the imitation learning to warm start our model training.

Recently, DRL has been used successfully to learn efficient policies in various fields. In the field of robot navigation, existing DRL methods learn policies from raw data (images or 2D laser data) directly \cite{tai2017virtual, long2018towards} or agent-level representation extracted from multiple sensors \cite{chen2017socially,chen2017decentralized,everett2018motion}. Although the raw data methods have the advantage of handing a variable number of agents directly, it is impossible to obtain a richer high-level representation of human intentions and interactions.

To address a variable number of humans in agent-level representation methods, Everett \textit{et al.} \cite{everett2018motion} introduced an LSTM module to convert the variable crowd to a fixed dimension input. In this method, the states of humans were input in descending order according to the distance between humans and robot, which is not always reasonable. Chen \textit{et al.} \cite{chen2019crowd} used local maps and self-attention module to encode the state of a variable crowed and capture the pairwise interactions among all agents that are better than the one-way interactions in LSTM. More recently, RGL \cite{chen2020relational} used Graph Convolutional Networks (GCNs) with self-attention mechanism to encode the interactions among all agents and modeled human motions by using a neural network that avoids the simplifying assumption for human motions modeling in \cite{chen2019crowd}. Building upon RGL, our method introduces the history information as important as the interactions. Furthermore, we employ Soft Actor-Critic (SAC) based on Actor-Critic framework in RL to avoid modeling the human motions and design a selection module after the policy network in SAC.

\subsection{Problem Formulation}

In this work, we consider a robot navigation task where the robot is required to move from the starting point to the goal on the plan $ \mathbb R^2 $, while avoiding the $n$ non-communicating other agents surrounding it.

This can be formulated as a sequential decision-making problem in a RL framework \cite{chen2017decentralized}. The states of an agent that can be observed by other agents include position $\mathbf P=\left[p_x,p_y\right]$, velocity $\mathbf V=\left[v_x,v_y\right]$, and radius $r$. The hidden states $\mathbf s^h$ of the robot include the goal position $\mathbf P_g$, preferred speed $\mathbf V_{pref}$. $\mathbf v_{i,t}^{his}$ denotes the history state composed of the observed states for the $i\text{-th}$ agent at time $t$, the script $i\in \{0,...,n\}$, where $i=0$ denotes the robot which is omitted. $\mathbf S_{i,t}^{his}$ denotes the history states set over continuous time ending at time $t$ for the $i\text{-th}$ agent. $\mathbf S_{H,t}^{his}=\left[\mathbf S_{1,t}^{his},\mathbf S_{2,t}^{his},...,\mathbf S_{n,t}^{his}\right]$ denotes the history states set for $n$ humans at time $t$. The joint state for the robot navigation is defined as $\mathbf S_{t}^{jn}=\left[\mathbf S_{t}^{his},\mathbf S_{H,t}^{his},\mathbf s^h\right]$. The action in a RL framework is defined as the displacement between the present position and the next position at time $t$, \textit{i.e.}, $\mathbf a_t=\left[\Delta p_x,\Delta p_y\right]$.

Accordingly, the robot observes a joint state $\mathbf S_{t}^{jn}$ in each time step $t$, then chooses an action $\mathbf a_t$ according to the joint state $\mathbf S_{t}^{jn}$, policy $\pi $, and selection module. After executing the command generated from the tracking layer based on the action, the robot will receive a reward $R(\mathbf S_{t}^{jn},\mathbf a_t)$ from the environment. Then $\mathbf S_{t}^{jn}$ will transfer to $\mathbf S_{t+1}^{jn}$.

SAC algorithm based on the maximum entropy is introduced in our work which aims to succeed at the task while acting as randomly as possible \cite{haarnoja2018soft}. That is, the maximum entropy will improve the exploration of the robot. In SAC, the policy is approximated by a neural network rather than obtained according to the value function, which avoids any arbitrary assumptions about state transition dynamics

The optimal policy ${{\pi }^{*}}$ in SAC is to maximize expected reward while also maximizing entropy:
\begin{subequations}
\begin{equation}
	\pi ^* =\underset{\pi }{\mathop{\arg \max }}\,\mathbb E_{(\mathbf S_{t}^{jn},\mathbf a_t)}\left[\sum\limits_{t}{R(\mathbf S_{t}^{jn},\mathbf a_t)}
	+H_e\right]
\end{equation}
\begin{equation}
	H_e=\alpha\mathcal H(\pi(\cdot|\mathbf S_{t}^{jn}))
\end{equation}
\end{subequations}
where $\alpha $ is a temperature hyper-parameter which determines the relative importance of the entropy term against the reward, $\mathcal H(\cdot)$ is an entropy term.

Inspired by \cite{brito2021go}, the reward function based on \cite{chen2017decentralized} adds progress reward:
\begin{subequations}
\begin{equation}
R(\mathbf S_{t}^{jn},\mathbf a_t)=
	\begin{cases}
	2&\text{if }\mathbf P_t=\mathbf P_g\\
	-0.4&\text{else if }d_\text{min}<R\\
	d_\text{min}-0.2&\text{else if }d_\text{min}<0.2+R\\
	i_t&\text{otherwise}
	\end{cases}
\end{equation}
\begin{equation}
	i_t=1.6*(||\mathbf P_t-\mathbf P_g||-||\mathbf P_{t+1}-\mathbf P_g||)
\end{equation}
\end{subequations} 
where $d_\text{min}$ is the minimum distance between robot and humans during the time period $\left[t-\Delta t,t\right]$, $R=r+r^i$,$r^i$ is radius of the $i\text{-th}$ human, $r$ is the radius of the robot, $0.2$ is the uncomfortable distance of humans, $i_t$ is the progress reward that encourages the robot to move towards the goal.

\section{Approach}
\begin{figure*}[!t]
	\centering
	\includegraphics[width=5in]{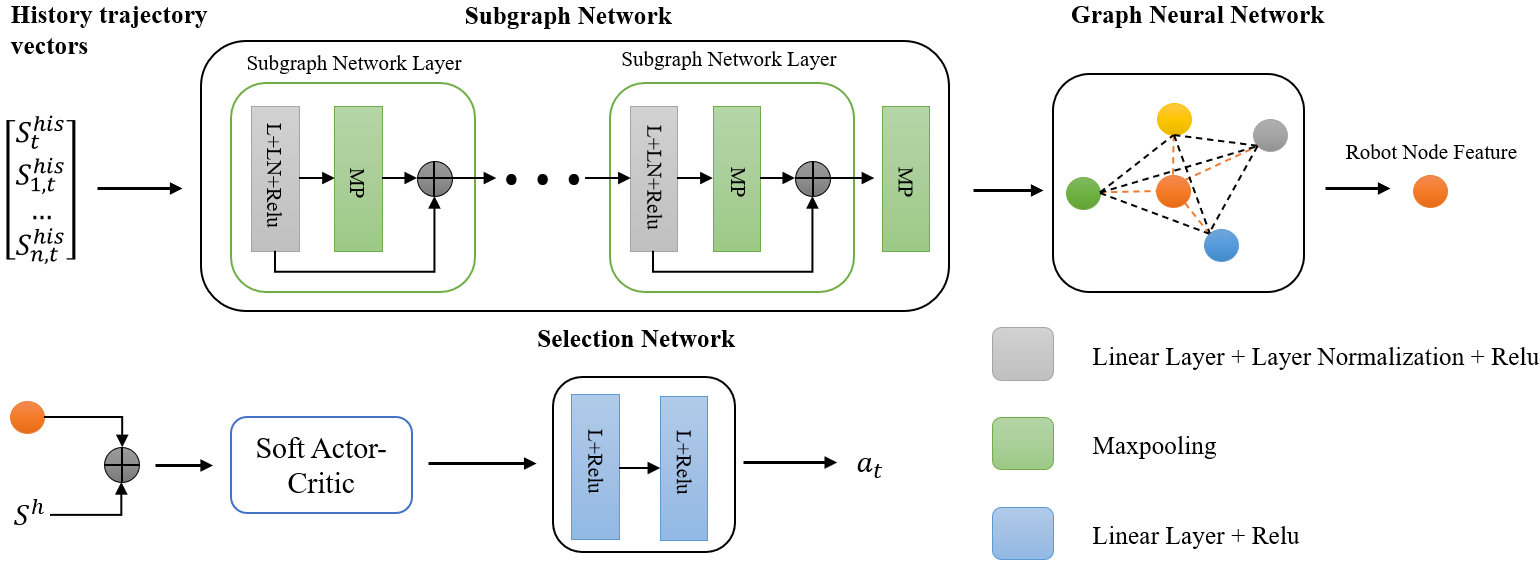}
	\caption{The proposed network structure including subgraph network, GNN, SAC network and selection network. The subgraph network consisting of three subgraph network layers aggregates the history information for each agent node. The GNN aggregates information about the crowd and outputs the robot node feature for SAC network to generate the policy. The selection module selects the best action from multiple sampling actions generated by policy. }
	\label{fig_2}
\end{figure*}
In this section, we first introduce how we vectorize history information of all agents and how we aggregate history information vectors belonging to the same agent by using a subgraph network. Then, the graph neural network will be described for modeling interactions among all agents. Finally, the selection module is shown. Fig. \ref{fig_2} shows the proposed network structure and the details are described in this section.

\subsection{Aggregating History Information}
\begin{figure}
	\centering
	\includegraphics[width=2.5in]{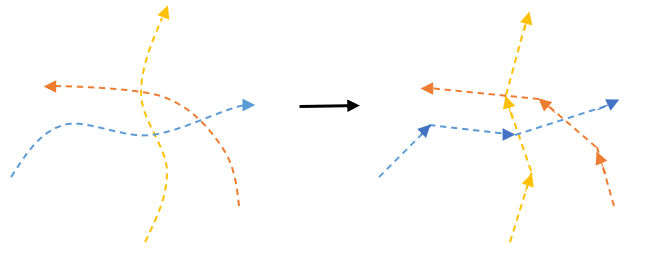}
	\caption{Agent trajectories vectorization. The left is the original trajectory, and the right is the vectorized trajectory.}
	\label{fig_3}
\end{figure}

In the robot navigation task, the trajectories of agents can be regarded as curves with respect to time  $t$. When key points in trajectories are sampled with a sufficiently small and fixed temporal interval $\Delta t$, the trajectories between two consecutive sampling points can be approximated at linear segments. Also, these linear segments that represent the agents moving from the last sampling point to the next sampling point contain direction information. Therefore, we present them as vectors, as depicted in Fig. \ref{fig_3}. In addition, the observable states of agents are added to vectors so that the robot could get more comprehensive information about humans and itself. More specifically, each vector $\mathbf v_{i,t}^{his}$  belonging to the $i\text{-th}$ agent at time $t$ is formulated as
\begin{equation}
\mathbf v_{i,t}^{his}=\left[ \mathbf P_{t-1}^i,\mathbf V_t^i,r_i,\mathbf P_t^i \right]
\end{equation}

In this work, the history information set includes trajectories of the last three time periods:
\begin{equation}
	\mathbf S_{i,t}^{his}=\left[ \mathbf v_{i,t-2}^{his},\mathbf v_{i.t-1}^{his},\mathbf v_{i.t}^{his} \right]
\end{equation}
subsequently, each history vector is transformed by using a multi-layer perceptron (MLP) that includes a linear layer, layer normalization and ReLU activation function:
\begin{equation}
	\mathbf A _t^i={{\phi }_{\alpha }}(\mathbf S_{i,t}^{his};\theta_\alpha)
\end{equation}
where $\theta_\alpha$ is the neural network weights shared by all agents, $\phi _\alpha $ is a MLP, $\mathbf A _t^i=\left[\alpha _{t-2}^{i},\alpha _{t-1}^{i},\alpha _{t}^{i}\right]$, is the projection corresponding to the history information set .

In order to deduce the intentions of agents from the history information so that the robot can avoid obstacles more effectively, we aggregate the $\mathbf A _t^i=\left[\alpha _{t-2}^{i},\alpha _{t-1}^{i},\alpha _{t}^{i}\right]$ into the feature $\mathbf b _t^i$ to extract historical correlation, and copy it into three, $\mathbf B _t^i=\left[b _{t}^{i},b _{t}^{i},b _{t}^{i}\right]$. Then, $\mathbf A_t^i$ are concatenated with $\mathbf B _t^i$ in feature dimension :
\begin{align}
	\mathbf B _t^i=\varphi (\mathbf A_i^t)\\
	\mathbf C_t^i=\mathbf A _t^i\oplus \mathbf B _t^i
\end{align}
where $\oplus $ is concatenation operation, $\varphi (\cdot )$ inculding copy operation and aggregation operation which is maxpooling operation in practice, $\mathbf C_t^i=\left[c_{t-2}^{i},c_{t-1}^{i},c_t^i\right]$, is the joint feature set, $c_t^i$ is the joint feature corresponding to the history information at time $t$.

Finally, to obtain agent level features, $\mathbf C_t^i$ is aggregated into $\mathbf h_t^i$ with aggregation operation:
\begin{equation}
	\mathbf h_t^i=\varphi (\mathbf C_t^i)
\end{equation}
where $\mathbf h_t^i$ is agent level features aggregating the history information.
\subsection{Modeling Interactions using a GNN}

In the robot navigation task, the robot, humans and their relationships can be modeled as a graph, so that the problem of the varying number of agents can be solved. The node of graph corresponds to the agent level features $\mathbf h_t^i$ of each agent, the edge of graph indicates how much attention one agent pays to another. Moreover, to learn a good representation of the interactions among all agents, the GNN is introduced in this work. 

An important component of the GNN is the adjacency matrix $\mathcal A$ which includes attention among all agents. It can be assigned in a certain way, such as using the spatial distances or using the temporal span. In our work, we use the trainable parameter based on self-attention mechanism to represent the adjacency matrix $\mathcal A$ , which can model more accurate high-order interactions.

The graph neural network layer is the key build of a GNN. It takes the original node features or the node features from the last layer as input and transforms them into output node features. Our graph neural network layer is implemented as a self-attention operation:
\begin{equation}
	\mathbf H_{t}^{l+1}=\sigma (\frac{\text{softmax}(\mathbf Q_{\mathbf H_t^l}\mathbf K^{\text{T}}_{\mathbf H_t^l})}{\sqrt{d_k}}\mathbf V_{\mathbf H_t^l})
\end{equation}
where $\mathbf H_t^l$ is the node features in layer $l$, $\mathbf Q,\mathbf K,\mathbf V$ are query matrix, key matrix, value matrix respectively, which are linear projections of $\mathbf H_t^l$, $d_k$ is the dimension of key matrix in feature dimension, $\sigma (\cdot )$ is the ReLU activation function. The $1/{\sqrt{d_k}}\;$ implements the scaled dot product term for numerical stability for attention \cite{yu2020spatio}. $\text{softmax}(\mathbf Q_{\mathbf H_t^l}\mathbf K^{\text{T}}_{\mathbf H_t^l})$ corresponds to the adjacency matrix $\mathcal A$. We introduce a linear neural network to represent linear projections, and use a single GNN layer in our implementation.

The robot feature is extracted from the GNN which includes history information and interaction information of all agents. In addition, we add hidden states of the robot itself, including the goal position $\mathbf P_g$ and the preferred speed $\mathbf V_{pref}$:
\begin{equation}
		\mathbf h_t^i=\mathbf h_t^i\oplus \left[\mathbf P_g,\mathbf V_{pref}\right]	
\end{equation}
Subsequently, the robot feature is fed into the policy network in SAC to generate the policy.

\subsection{Selection Module}
In SAC algorithm, policy is stochastic which makes action performed by the robot in a given state unfixed. 
Additionally, the maximum entropy in SAC also guarantees exploration \cite{haarnoja2018soft}. Therefore, the robot has multiple plausible paths in a fixed policy, which can solve the freezing robot problem to certain extent. Also, that is why we use SAC in this work.

However, the limitation of stochastic policy is that the probability of unreliable actions still exists in the inference phase. Moreover, the reliable actions are also different. Some actions are more aggressive, others are more conservative.

In order for the robot to navigate through crowded human environments in a stochastic policy, and satisfy the task requirements, the selection module is designed after the policy network. The reasons for it are that selecting action adaptively from the action set according to the current situation and reducing the probability of unreliable actions.

In our work, we sample multiple next position points from action distribution in policy network firstly. Then, we develop a selection network to estimate the Q-value of sampling points. Finally, we select a sampling point according to Q-value:
\begin{equation}
	\mathbf a_t=\arg {{\max }_{\mathbf a_{t}^{k}\in \{\mathbf a_{t}^{1},\ldots ,\mathbf a_{t}^{m}\}}}Q(\mathbf h_t^i,\mathbf a_t^k)
\end{equation}
where $m$ is the number of sampling points, $a_{t}^{k}$ is the action sampled from the distribution.

The above operation is similar to Deep Q-learning (DQN) that selects the action with the maximum Q-value from action space. However, the policy generated from the selection module is not deterministic. In our work, the action space will change with the change of sampling points because of the distribution, which is different from the fixed action space in DQN.

The selection module plays an auxiliary role in the selection of the robot action, which further selects action sampled from the action space. Early in the training phase, the action performed by the robot mainly depends on the distribution generated from the policy network in SAC. Later in the training phase, it mainly depends on the selection module after the policy network converges. 

\subsection{SH-SACGNN}

 We refer to our full method with Selection module and History information based on SAC and GNN as SH-SACGNN. The training algorithm is shown in Algorithm \ref{algo_disjdecomp}, where ${\theta}_{sub}$ denotes the subgraph network paremeter, ${\theta}_{gnn}$ denotes the GNN parameter, ${\theta}_{\pi}$ denotes the policy network parameter, ${\theta}_{select}$ denotes the selection network parameter.
  \IncMargin{1em}
 \begin{algorithm}
 	\SetKwData{Left}{left}\SetKwData{This}{this}\SetKwData{Up}{up}
 	\SetKwFunction{Union}{Union}\SetKwFunction{FindCompress}{FindCompress}
 	\SetKwInOut{Input}{input}\SetKwInOut{Output}{output}
 	\SetAlgoNoLine 
 	\BlankLine
 	\text{Initialize all networks with demonstration D }\\
 	\text{Initialize target Q-value network with Q-value network}\\
 	\text{Initialize replay buffer B $\gets$ D}\\
 	\For{episode = 1 to $N_{eps}$ }{
 		\text{Initialize the join state $\mathbf S_{0}^{jn}$} through vectorized representation \\
 		\While{not done}{
 			\text{Vectorize states of all agents}\\
 			\For{i = 1 to $m$}{
 			$\text{Sample action } \mathbf a_{t}^{i} \text{ from F}_p(\mathbf S_{t}^{jn};{\theta}_{sub},{\theta}_{gnn},{\theta}_{\pi})$\\
 			$\text{Action set } \mathbf {as}_ t\gets  \mathbf a_{t}^{i}$\\
 			}
 			
 			$\mathbf a_t=\text{F}_s(\mathbf {as}_ t;{\theta}_{select})$\\
 			$\{\mathbf S_{t+1}^{jn},R(\mathbf S_{t}^{jn},\mathbf a_t),\text{done }\} = \text{step} (\mathbf S_{t}^{jn},\mathbf a_t)$\\
 			$\text{Store B}\gets (\mathbf S_{t}^{jn},\mathbf a_t,R(\mathbf S_{t}^{jn},\mathbf a_t),\mathbf S_{t+1}^{jn},\text{done})$\\
 			\text{Training with data from B} \\
 			\text{Update target Q-value network}\\
 		}
 	}	
 	\caption{SH-SACGNN}\label{algo_disjdecomp}
 \end{algorithm}\DecMargin{1em} 
 
 Training consisted of two stages: imitation learning and reinforcement learning. In the imitation learning stage, the replay buffer B is initialized with demonstration D generated from ORCA policy and all networks are initialized by supervised learning. Imitation learning is a important stage for model initialization, because it can greatly speed up the training in RL. In the reinforcement learning stage, for each step, the state of all agents are be vectorized first. Then, we get action set $\mathbf {as}_{t}$ including $m$ actions from the policy function $\text{F}_p(\cdot)$ according to the joint state $\mathbf S_{t}^{jn}$. Furthermore, the best action $\mathbf {a}_ t$ in action set is selected by the selection function $\text{F}_s(\cdot)$. Finally, the action is input into the tracking layer and the robot executes command. Selection function is updated with temporal-difference error, and the SAC  is update as described in \cite{haarnoja2018soft}.

\section{Experiments}
In this section, the setting of simulation environment, the implementation details of neural networks and hyper-parameters are described first. Subsequently, we evaluate the performance of our method by providing ablation studies and comparing it against the existing methods, ORCA\cite{van2011reciprocal} and RGL \cite{chen2020relational}. Besides our full method SH-SACGNN, the method without selection module is referred as H-SACGNN, the method without history information and selection module is referred as SACGNN. The methods are trained and tested with 500 random test cases in the environments with 5 humans and 10 humans. Finally, a implementation of our method in the real world is shown in the last subsection.

\subsection{Simulation Setup}
We use the CrowdNav simulation environment proposed by Chen \cite{chen2019crowd}. In common with RGL, the circle crossing scenario with radius $4m$ is used for our experiments. The initial positions of agents are randomly positioned on the circumference with random perturbation added to their $x,y$ coordinates within $0.5m$ .

The radius of agents is $0.3m$, which corresponds to radius of humans in the real situation, and the goals are the symmetry points of initial position about the center of the circle. The maximum velocity for the robot $\mathbf v_{pref}$ is $1{m}/{s}$. Considering that the robot navigates in a social etiquette way, we set the robot to be invisible to humans, that is, the human only needs to consider other humans when moving. The robot has to actively avoid humans in this situation. In this work, we assume holonomic kinematics for the robot, \textit{i.e.}, it can move in any direction. The robot is controlled by our method, and humans are controlled by ORCA.
\subsection{Implementation Details}
In subgraph network, we use three subgraph network layers, and the hidden dimension of $\phi (\cdot )$ is $64$ . The hidden dimension of the linear projection in the GNN for $\mathbf Q,\mathbf K,\mathbf V$ is $128$. The hidden units of critic network, policy network, selection network are $(256,\text{ }256,\text{ }256)$, $(256,\text{ }256,\text{ }256)$, $(256,\text{ }256)$ respectively. All trainable parameters are trained with a batch size of $256$ using Adam \cite{kingma2014adam}. The size of replay buffer is set to be $400k$. The discount factor $\gamma $ is set to be $0.99$. The episodes for imitation learning and reinforcement learning are set to be $2k$, $15k$ respectively. The learning rates for imitation learning and reinforcement learning are set to be ${{10}^{-3}}$, $3\times {{10}^{-4}}$ respectively. The temperature parameter $\alpha $ is set to be 0.05. The temporal interval $\Delta t$ is set to be $0.25$ seconds.In selection module, the number $m$ of input actions sampled from policy network is set to be $4$. In addition, in simulation environment, we model robot motion with a linear model in the tracking layer, \textit{i.e.}, $\mathbf V_t=\Delta \mathbf P_t/t$.

\begin{table}
	\begin{center}
		\caption{Quantitative results in the environments with 5 humans. The metrics includes success rate, collision rate, timeout rate and average navigation time in 500 tests.}
		\label{tab1}
		\begin{tabular}{ c  c  c  c  c }
			\hline
			Methods & success(\%) & collision(\%) & timeout(\%)& time(s) \\
			\hline
			ORCA \cite{van2011reciprocal}&43.4&56.6&\textbf{0.0}&10.91\\
			\hline
			RGL\cite{chen2020relational}&93.0&4.0&3.0&\textbf{9.98}\\ 
			\hline
			SACGNN(our)&97.8&1.8&0.4&10.51\\
			\hline 
			H-SACGNN(our)&98.4&1.6&\textbf{0.0}&11.49\\
			\hline 
			SH-SACGNN(our)&\textbf{99.4}&\textbf{0.6}&\textbf{0.0}&10.98\\
			\hline 
		\end{tabular}
	\end{center}
\end{table}

\begin{table}
	\begin{center}
		\caption{Quantitative results in the environments with 10 humans.}
		\label{tab2}
		\begin{tabular}{ c  c  c  c  c }
			\hline
			Methods & success(\%) & collision(\%) & timeout(\%)& time(s) \\
			\hline
			RGL\cite{chen2020relational}&88.6&5.6&5.8&12.55\\ 
			\hline
			SACGNN(our)&95.1&3.9&1.0&12.51\\
			\hline 
			H-SACGNN(our)&96.8&3.0&0.2&12.82\\
			\hline 
			SH-SACGNN(our)&\textbf{98.4}&\textbf{1.6}&\textbf{0.0}&\textbf{12.15}\\
			\hline 
		\end{tabular}
	\end{center}
\end{table}

\subsection{Quantitative Evaluation}
\begin{figure}[!t]
	\centering
	\subfloat[ {\small RGL}]{\includegraphics[width=1.6in]{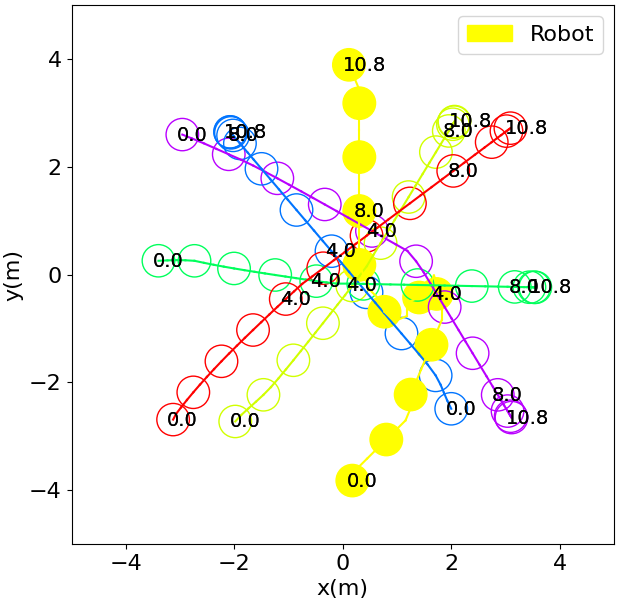}%
	\label{fig4a}}
	\hfil
	\subfloat[\small SACGNN]{\includegraphics[width=1.6in]{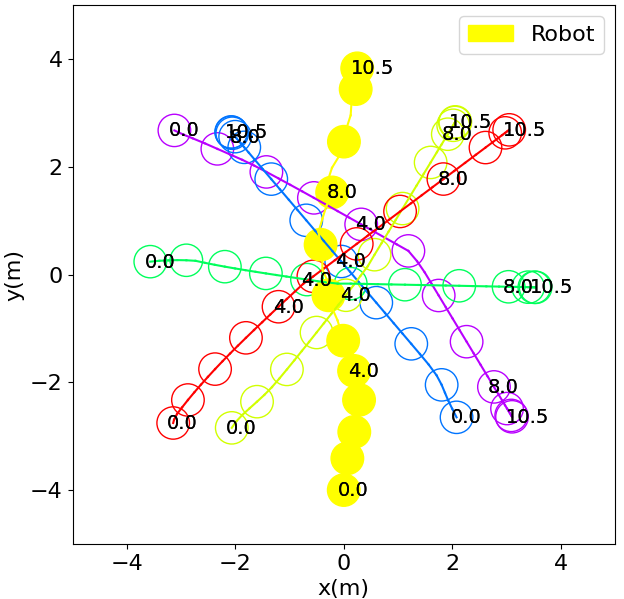}%
	\label{fig4b}}
	
	\subfloat[\small H-SACNGNN]{\includegraphics[width=1.6in]{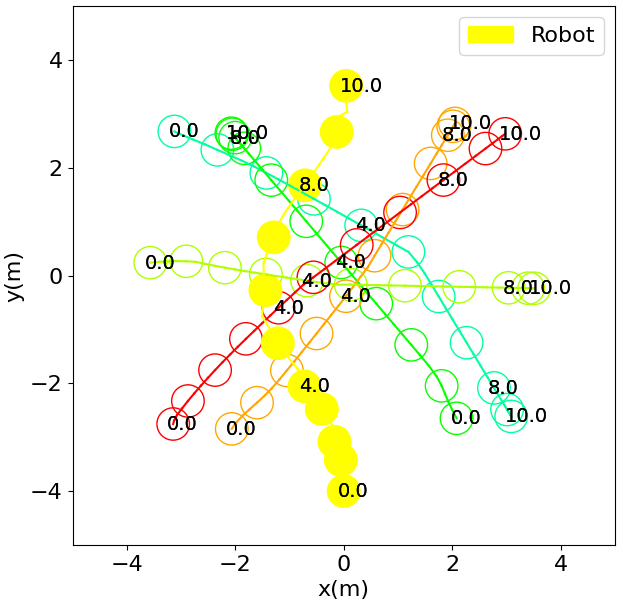}%
	\label{fig4c}}
	\hfil
	\subfloat[\small SH-SACGNN]{\includegraphics[width=1.6in]{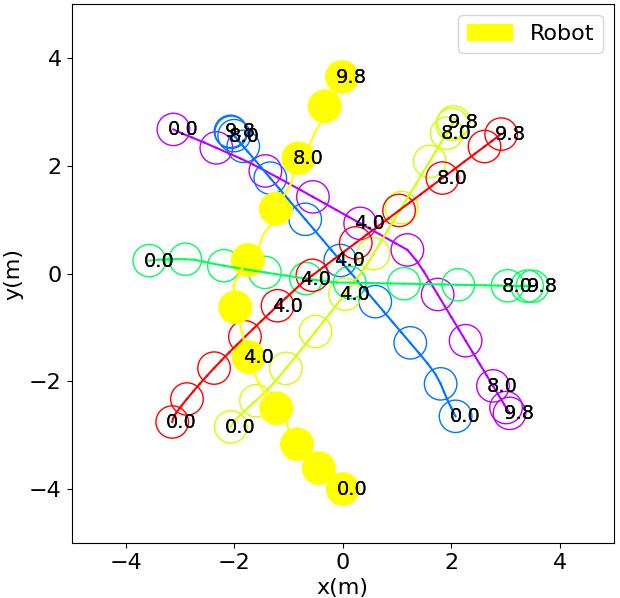}%
	\label{fig4d}}
	\caption{Trajectory comparison. The highlighted yellow lines show the robot trajectories. The other colors show the human trajectories.}
	\label{fig4}
\end{figure}
\begin{figure}
	\centering
	\subfloat[]{\includegraphics[width=1.6in]{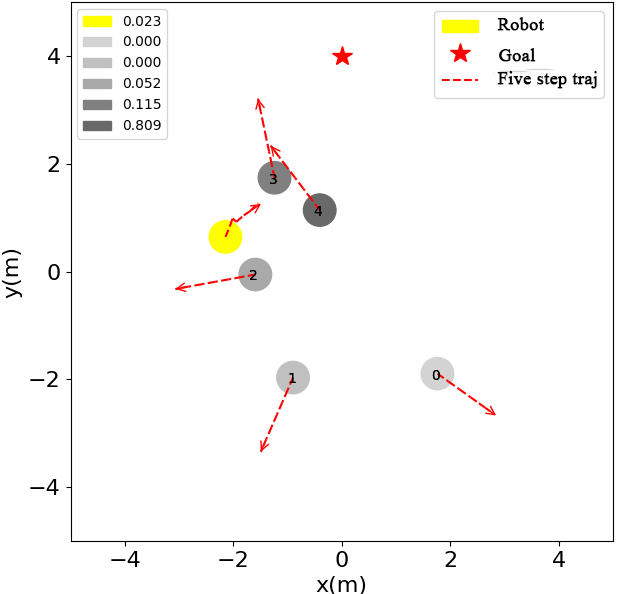}%
		\label{fig5a}}
	\hfil
	\subfloat[]{\includegraphics[width=1.6in]{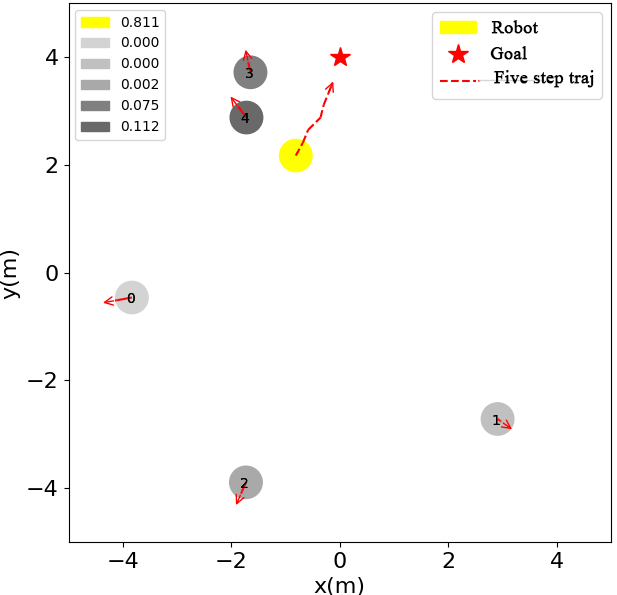}%
		\label{fig5b}}
	\caption{Attention weights from our method. The darker the color of the circle representing humans, the greater the weights, and the specific score is shown in the upper left corner of figure.}
	\label{fig5}
\end{figure}
\begin{figure}
	\centering
	\subfloat[]{\includegraphics[width=1.6in]{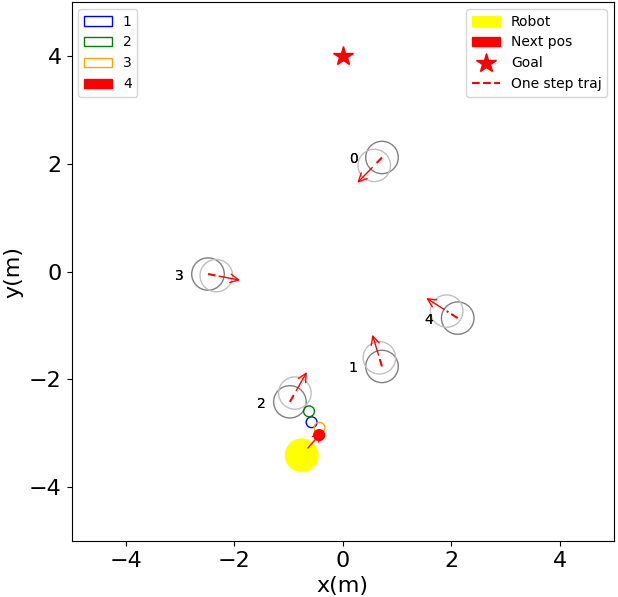}%
		\label{fg6a}}
	\hfil
	\subfloat[]{\includegraphics[width=1.6in]{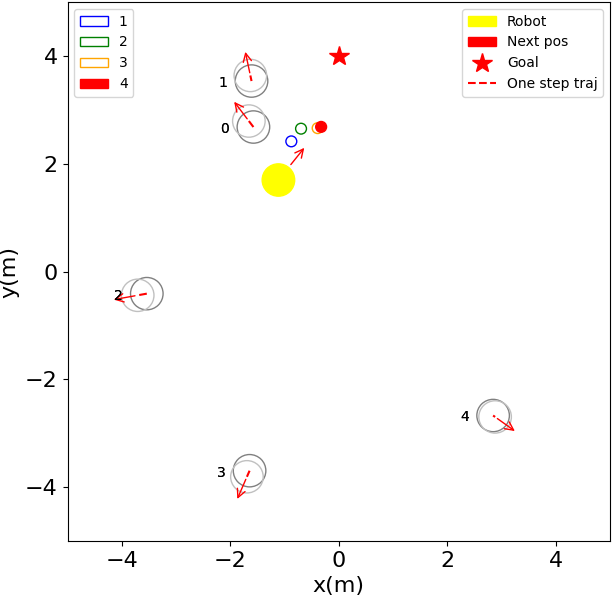}%
		\label{fg6b}}
	\caption{The little circles indicate four next position points sampled from the policy network, where the red circle indicates the actual next position point of the robot selected by the selection module. The red dashed line indicates the future trajectory in next one steps. The radius of the next position points is $1m$.}
	\label{fig6}
\end{figure}
\subsubsection{Comparing with Other Methods}
Table \ref{tab1} reports the success rate, collision rate, timeout rate and the average navigation time of successful tests in the environments with 5 humans. As expected, the ORCA has a very low success rate, only $43.4\%$, due to the violation of the reciprocal assumption in the invisible setting. The state-of-the-art method RGL has a higher success rate of $93\%$. Compared with RGL, our method SH-SACGNN achieves the highest success rate, which is $99.4\%$. Apart from the success rate, our method reduces the timeout rate from $3\%$ to $0\%$ and reduces the collision rate from $4\%$ to $0.6\%$. Although the average navigation time of our method is about $1s$ slower than that of RGL, it is acceptable. In the robot navigation task, the first consideration is to avoid collisions between the robot and humans or other objects. Because the collision will bring some bad consequences, including human injury, robot damage and so on. In contrast, the navigation time only represents the efficiency of the robot to complete the task. For this phenomenon, we think that our model completed the complicated tests that RGL can not, resulting in a long navigation time in these tests, which makes the total average navigation time longer in successful tests.

In the environments with 10 humans, as shown in Table \ref{tab2}, the timeout rate of RGL has risen. However, the timeout rate of our method remains zero, and the success rate is much higher than that in RGL, about $10\%$ higher. From the environments with 5 humans to the environments with 10, the success rate of RGL drops by $4.4\%$. In comparison, our method only drops $1\%$. These results show our method SH-SACGNN including history information and selection module is safer than RGL, and can suppress the performance deterioration as the crowd grows to certain extent.
\subsubsection{Ablation Studies}To evaluate the benefits of adding history information, we compare SACGNN and H-SACGNN. It can be found that although the addition of history information in the environments with 5 humans (Table \ref{tab1}) only slightly improves success rate, about $0.6\%$ and increases average navigation time, this improvement is enhanced in the environments with 10 (Table \ref{tab2}) humans. Also, the gap in average navigation time reduce from $1s$ to $0.3s$. The results suggest that history information improves the performance of robot in terms of success rate, collision rate and timeout rate, especially in crowded environments. In other words, adding history information improves the ability of the robot to anticipate the evolution of crowds by predicting future trajectories implicitly so that it can recognize a feasible path.

We further evaluate the benefits of selection module by comparing H-SACGN and SH-SACGNN. In the environments with 5 humans (Table \ref{tab1}), we observe that the introduction of the selection module improves all metrics compared to H-SACGNN. Moreover, this improvement is more obvious in the environments with 10 humans (Table \ref{tab2}). The results show that the selection module has a positive effect in the selection of actions, including safety and rapidity.

Besides, comparing the SACGNN and RGL, it is apparent that no matter in which environments in  our work, the timeout rate is reduced in SACGNN, which proves that SAC can reduce the robot freezing problem to certain extent.

\begin{figure}[!t]
	\centering
    \includegraphics[width=1.5in]{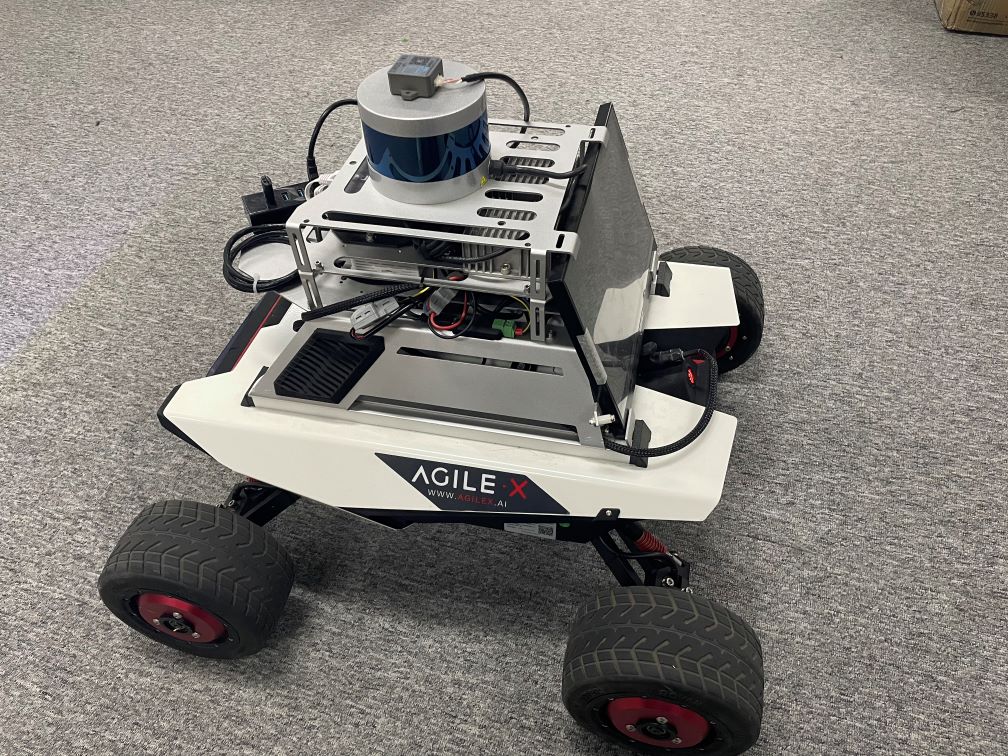}%
	\caption{The SCOUT MINI robot we used with a laptop and a velodyne-16 LIDAR in the real-world environments.}
	\label{fig7}
\end{figure}

\begin{figure}[!t]
    \centering
	\includegraphics[width=1.5in]{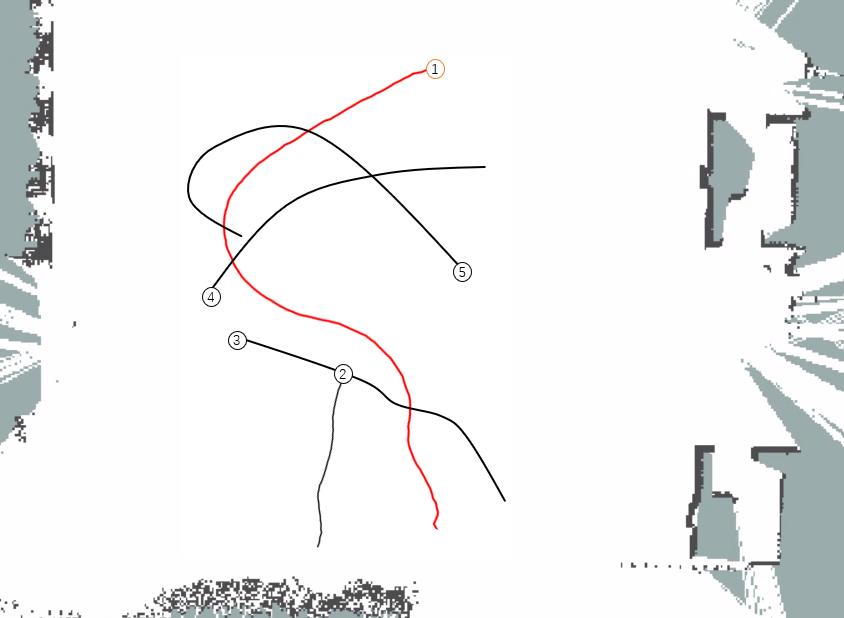}%
	\caption{Trajectories are presented in a real-world demonstration via RVIZ. The read line indicates the trajectory of the robot and the black lines indicate the trajectories of humans. The circles with a number indicate their goal.}
	\label{fig8}
\end{figure}
\subsection{Qualitative Evaluation}
\subsubsection{Trajectory Comparison}
Fig. \ref{fig4} shows example trajectories generated by RGL, SACGCN, H-SACGNN and SH-SACGNN in the environments with 5 humans. For a clear comparison, we choose the same test case where the trajectories of humans are identical. The RGL tries to choose the right path at first to move fast, but is threatened by the purple human and has to turn back to choose the middle path. Although the SACGNN hesitates at first, it keeps taking the middle path, ending up with a shorter navigation time.

In contrast, the H-SACGNN including history information hesitates at first but then finds a shorter path to the goal. The explanation is that the robot can predict the future trajectories of humans from history information. In this scene, humans move from the left to the right resulting in a lot of space on the left. Based on this, The robot chooses the left path to reach the destination quickly. As shown in Fig. \ref{fig4d}, the SH-SACGNN further optimizes the trajectory of the robot, which reduces the navigation time and the hesitation steps at first.

\subsubsection{Visualizing Attention Weights}
\begin{figure}[!t]
	\centering
	\subfloat[]{\includegraphics[width=1.7in]{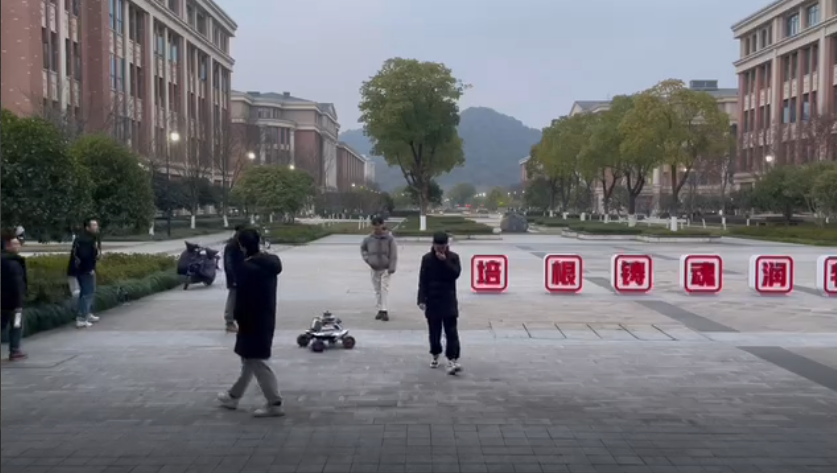}%
		\label{fg9a}}
	\hfil
	\subfloat[]{\includegraphics[width=1.7in]{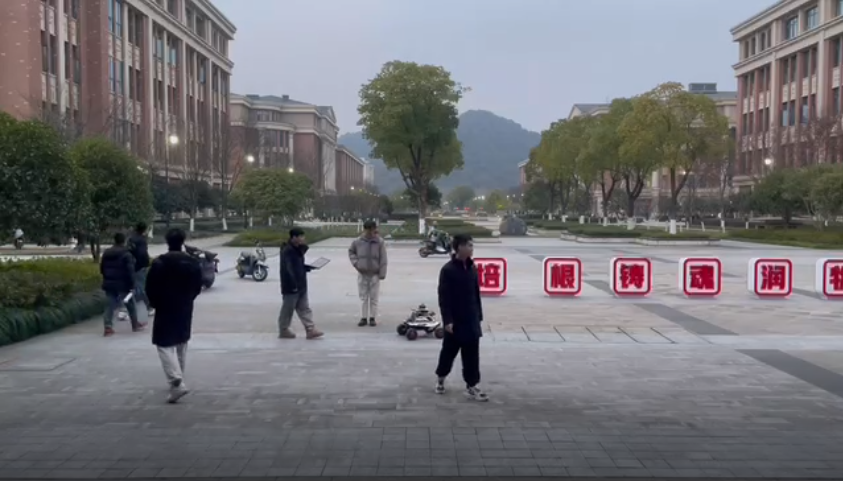}%
		\label{fg9b}}

	\subfloat[]{\includegraphics[width=1.7in]{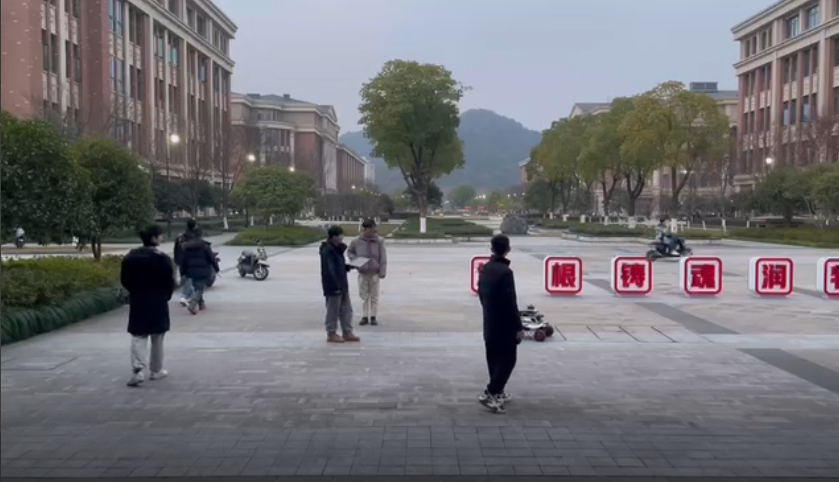}%
		\label{fg9c}}
	\hfil
	\subfloat[]{\includegraphics[width=1.7in]{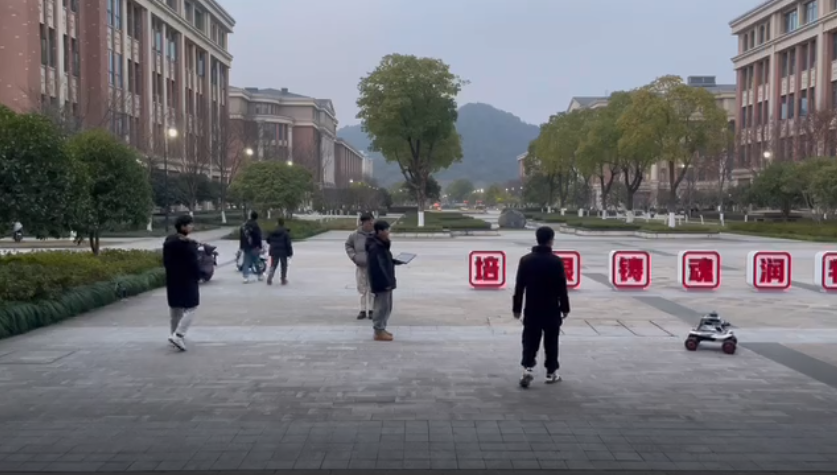}%
		\label{fg9d}}
	\hfil
	
	\caption{The four chronological moments in the real-world environment.}
	\label{fig9}
\end{figure}
We further visualize the attention weights inferred by our full method. As shown in Fig. \ref{fig5a}, the lowest attention score is assigned to Human $0$ and Human $1$ who have the largest distance to the robot. The Human $2$ closest to the robot receives a low score, as it is moving in the opposite direction of the robot. The robot pays more attention to Human $3$ and Human $4$. However, the attention for Human $4$ is far greater than that for Human $3$ which is closer to the robot. To analyze this phenomenon, we plot the future trajectories in the next five steps. It can be seen that although the robot moves towards Human $3$ at the current moment, the influence of Human $4$ on the robot becomes more and more in the next five steps. The result demonstrates that our interaction module including history information used to predict future trajectories implicitly has a good ability to reason the relative importance and confirms the attention depends not only on the current state but also on the future state.

Compared to the robot in Fig. \ref{fig5a}, the robot in Fig. \ref{fig5b} pays more attention to itself rather than humans. In the first situation, the more urgent thing for the robot is to avoid humans, while the robot hopes to reach the goal quickly in the second situation.

\subsubsection{Visualizing Sampling Points}

In addition, We investigate the effectiveness of selection module through qualitative analysis. The Fig. \ref{fg6a} shows that not all next position points sampled from the policy network in the inference phase satisfy the task requirements. The position point $1$ and position point $2$ will collide with Human $2$ in the next step, and the position point $3$ is within the uncomfortable range of Human $2$. Although the selected position point $4$ is the farthest away from the goal, it is the safest.

In contrast, as shown in Fig. \ref{fg6b}, all next position points sampled from the policy network satisfy the task requirements. In this situation, our selection module selected the position point $4$ as the next position point of the robot because it is closest to the goal, which will reduce navigation time. The results prove that our selection module has the ability to select the next position point adaptively and further improve output of the policy network.

\subsection{Real-world Experiments}
In addition to simulation experiments, we examine the trained policy in the real-world experiments on a SCOUT MINI robot equipped with a velodyne-16 LIDAR (Fig. \ref{fig7}). Based on the 2D raster map built with Gmapping algorithm, the 2D positions of robot and humans are obtained by using Adaptive Monte Carlo algorithm in the global coordinate system, where the origin of the coordinates is the starting point of the robot, and the x-axis is the starting direction of the robot. we then utilize an Kalman filter and DBSCAN clustering algorithm to track the humans and to compute human velocities. Finally, PID controller is used to make the robot to track the next position point in the real-world experiments. 

The trajectories in a real-world demonstration is shown in Fig. \ref{fig8}. In order to avoid collision with human $3$, the robot moves to the left firstly. As the human $2$ gets closer, the robot has to move slightly to the right. Then, based on predictions of human behavior which human $4$ moves to the left and human $5$ moves from top left to bottom right, the robot continues to move to its goal from the left path. This demonstration shows the feasibility of our method in the real-world environments. The Fig. \ref{fig9} shows the four moments in a real-world demonstration and the video demo can be found at \href{https://www.youtube.com/watch?v=0-FLAEh3iwY}{https://www.youtube.com/watch?v=0-FLAEh3iwY}.

\section{conclusion}
In this paper, we present a new collision avoidance approach for robot navigation tasks, which enables the robot to navigate through a crowd safely. There are two main contributions in this work. The first is the introduction of history information before modeling interactions by using a GNN. The history information improves the ability of the robot to predict the future trajectories of humans, which makes it more reasonable for the robot to pay attention to humans, resulting in a high success rate. The second is the design of a selection module after the policy network in RL framework. The purposes of the selection module are to reduce the unreliable actions sampled from the policy network and to select action adaptively according to the current situation. The simulation experiments demonstrated the superiority of the proposed approach over the state-of-the-art methods. Furthermore, we show the feasibility of our method in the real world.

\bibliographystyle{IEEEtran}
\bibliography{ref}{}

\begin{thebibliography}{10}
\providecommand{\url}[1]{#1}
\csname url@samestyle\endcsname
\providecommand{\newblock}{\relax}
\providecommand{\bibinfo}[2]{#2}
\providecommand{\BIBentrySTDinterwordspacing}{\spaceskip=0pt\relax}
\providecommand{\BIBentryALTinterwordstretchfactor}{4}
\providecommand{\BIBentryALTinterwordspacing}{\spaceskip=\fontdimen2\font plus
\BIBentryALTinterwordstretchfactor\fontdimen3\font minus
  \fontdimen4\font\relax}
\providecommand{\BIBforeignlanguage}[2]{{%
\expandafter\ifx\csname l@#1\endcsname\relax
\typeout{** WARNING: IEEEtran.bst: No hyphenation pattern has been}%
\typeout{** loaded for the language `#1'. Using the pattern for}%
\typeout{** the default language instead.}%
\else
\language=\csname l@#1\endcsname
\fi
#2}}
\providecommand{\BIBdecl}{\relax}
\BIBdecl

\bibitem{kayukawa2019bbeep}
S.~Kayukawa, K.~Higuchi, J.~Guerreiro, S.~Morishima, Y.~Sato, K.~Kitani, and
  C.~Asakawa, ``Bbeep: A sonic collision avoidance system for blind travellers
  and nearby pedestrians,'' in \emph{Proceedings of the 2019 CHI Conference on
  Human Factors in Computing Systems}, 2019, pp. 1--12.

\bibitem{chen2019crowd}
C.~Chen, Y.~Liu, S.~Kreiss, and A.~Alahi, ``Crowd-robot interaction:
  Crowd-aware robot navigation with attention-based deep reinforcement
  learning,'' in \emph{2019 International Conference on Robotics and Automation
  (ICRA)}.\hskip 1em plus 0.5em minus 0.4em\relax IEEE, 2019, pp. 6015--6022.

\bibitem{rudenko2020human}
A.~Rudenko, L.~Palmieri, M.~Herman, K.~M. Kitani, D.~M. Gavrila, and K.~O.
  Arras, ``Human motion trajectory prediction: A survey,'' \emph{The
  International Journal of Robotics Research}, vol.~39, no.~8, pp. 895--935,
  2020.

\bibitem{fong2003survey}
T.~Fong, I.~Nourbakhsh, and K.~Dautenhahn, ``A survey of socially interactive
  robots,'' \emph{Robotics and autonomous systems}, vol.~42, no. 3-4, pp.
  143--166, 2003.

\bibitem{kruse2013human}
T.~Kruse, A.~K. Pandey, R.~Alami, and A.~Kirsch, ``Human-aware robot
  navigation: A survey,'' \emph{Robotics and Autonomous Systems}, vol.~61,
  no.~12, pp. 1726--1743, 2013.

\bibitem{fox1997dynamic}
D.~Fox, W.~Burgard, and S.~Thrun, ``The dynamic window approach to collision
  avoidance,'' \emph{IEEE Robotics \& Automation Magazine}, vol.~4, no.~1, pp.
  23--33, 1997.

\bibitem{van2011reciprocal}
J.~Van Den~Berg, S.~J. Guy, M.~Lin, and D.~Manocha, ``Reciprocal n-body
  collision avoidance,'' in \emph{Robotics research}.\hskip 1em plus 0.5em
  minus 0.4em\relax Springer, 2011, pp. 3--19.

\bibitem{van2011re}
J.~Van Den~Berg, J.~Snape, S.~J. Guy, and D.~Manocha, ``Reciprocal collision
  avoidance with acceleration-velocity obstacles,'' in \emph{2011 IEEE
  International Conference on Robotics and Automation}.\hskip 1em plus 0.5em
  minus 0.4em\relax IEEE, 2011, pp. 3475--3482.

\bibitem{aoude2013probabilistically}
G.~S. Aoude, B.~D. Luders, J.~M. Joseph, N.~Roy, and J.~P. How,
  ``Probabilistically safe motion planning to avoid dynamic obstacles with
  uncertain motion patterns,'' \emph{Autonomous Robots}, vol.~35, no.~1, pp.
  51--76, 2013.

\bibitem{unhelkar2015human}
V.~V. Unhelkar, C.~P{\'e}rez-D'Arpino, L.~Stirling, and J.~A. Shah,
  ``Human-robot co-navigation using anticipatory indicators of human walking
  motion,'' in \emph{2015 IEEE International Conference on Robotics and
  Automation (ICRA)}.\hskip 1em plus 0.5em minus 0.4em\relax IEEE, 2015, pp.
  6183--6190.

\bibitem{kim2013predicting}
S.~Kim, S.~J. Guy, W.~Liu, R.~W. Lau, M.~C. Lin, and D.~Manocha, ``Predicting
  pedestrian trajectories using velocity-space reasoning,'' in
  \emph{Algorithmic foundations of robotics X}.\hskip 1em plus 0.5em minus
  0.4em\relax Springer, 2013, pp. 609--623.

\bibitem{trautman2010unfreezing}
P.~Trautman and A.~Krause, ``Unfreezing the robot: Navigation in dense,
  interacting crowds,'' in \emph{2010 IEEE/RSJ International Conference on
  Intelligent Robots and Systems}.\hskip 1em plus 0.5em minus 0.4em\relax IEEE,
  2010, pp. 797--803.

\bibitem{chen2020robot}
Y.~Chen, C.~Liu, B.~E. Shi, and M.~Liu, ``Robot navigation in crowds by graph
  convolutional networks with attention learned from human gaze,'' \emph{IEEE
  Robotics and Automation Letters}, vol.~5, no.~2, pp. 2754--2761, 2020.

\bibitem{helbing1995social}
D.~Helbing and P.~Molnar, ``Social force model for pedestrian dynamics,''
  \emph{Physical review E}, vol.~51, no.~5, p. 4282, 1995.

\bibitem{ferrer2013robot}
G.~Ferrer, A.~Garrell, and A.~Sanfeliu, ``Robot companion: A social-force based
  approach with human awareness-navigation in crowded environments,'' in
  \emph{2013 IEEE/RSJ International Conference on Intelligent Robots and
  Systems}.\hskip 1em plus 0.5em minus 0.4em\relax IEEE, 2013, pp. 1688--1694.

\bibitem{mehta2016autonomous}
D.~Mehta, G.~Ferrer, and E.~Olson, ``Autonomous navigation in dynamic social
  environments using multi-policy decision making,'' in \emph{2016 IEEE/RSJ
  International Conference on Intelligent Robots and Systems (IROS)}.\hskip 1em
  plus 0.5em minus 0.4em\relax IEEE, 2016, pp. 1190--1197.

\bibitem{mavrogiannis2019multi}
C.~I. Mavrogiannis and R.~A. Knepper, ``Multi-agent path topology in support of
  socially competent navigation planning,'' \emph{The International Journal of
  Robotics Research}, vol.~38, no. 2-3, pp. 338--356, 2019.

\bibitem{alahi2016social}
A.~Alahi, K.~Goel, V.~Ramanathan, A.~Robicquet, L.~Fei-Fei, and S.~Savarese,
  ``Social lstm: Human trajectory prediction in crowded spaces,'' in
  \emph{Proceedings of the IEEE conference on computer vision and pattern
  recognition}, 2016, pp. 961--971.

\bibitem{gupta2018social}
A.~Gupta, J.~Johnson, L.~Fei-Fei, S.~Savarese, and A.~Alahi, ``Social gan:
  Socially acceptable trajectories with generative adversarial networks,'' in
  \emph{Proceedings of the IEEE Conference on Computer Vision and Pattern
  Recognition}, 2018, pp. 2255--2264.

\bibitem{chen2017socially}
Y.~F. Chen, M.~Everett, M.~Liu, and J.~P. How, ``Socially aware motion planning
  with deep reinforcement learning,'' in \emph{2017 IEEE/RSJ International
  Conference on Intelligent Robots and Systems (IROS)}.\hskip 1em plus 0.5em
  minus 0.4em\relax IEEE, 2017, pp. 1343--1350.

\bibitem{chen2017decentralized}
Y.~F. Chen, M.~Liu, M.~Everett, and J.~P. How, ``Decentralized
  non-communicating multiagent collision avoidance with deep reinforcement
  learning,'' in \emph{2017 IEEE international conference on robotics and
  automation (ICRA)}.\hskip 1em plus 0.5em minus 0.4em\relax IEEE, 2017, pp.
  285--292.

\bibitem{everett2018motion}
M.~Everett, Y.~F. Chen, and J.~P. How, ``Motion planning among dynamic,
  decision-making agents with deep reinforcement learning,'' in \emph{2018
  IEEE/RSJ International Conference on Intelligent Robots and Systems
  (IROS)}.\hskip 1em plus 0.5em minus 0.4em\relax IEEE, 2018, pp. 3052--3059.

\bibitem{chen2020relational}
C.~Chen, S.~Hu, P.~Nikdel, G.~Mori, and M.~Savva, ``Relational graph learning
  for crowd navigation,'' in \emph{2020 IEEE/RSJ International Conference on
  Intelligent Robots and Systems (IROS)}.\hskip 1em plus 0.5em minus
  0.4em\relax IEEE, 2020, pp. 10\,007--10\,013.

\bibitem{gao2020vectornet}
J.~Gao, C.~Sun, H.~Zhao, Y.~Shen, D.~Anguelov, C.~Li, and C.~Schmid,
  ``Vectornet: Encoding hd maps and agent dynamics from vectorized
  representation,'' in \emph{Proceedings of the IEEE/CVF Conference on Computer
  Vision and Pattern Recognition}, 2020, pp. 11\,525--11\,533.

\bibitem{van2008reciprocal}
J.~Van~den Berg, M.~Lin, and D.~Manocha, ``Reciprocal velocity obstacles for
  real-time multi-agent navigation,'' in \emph{2008 IEEE International
  Conference on Robotics and Automation}.\hskip 1em plus 0.5em minus
  0.4em\relax IEEE, 2008, pp. 1928--1935.

\bibitem{phillips2011sipp}
M.~Phillips and M.~Likhachev, ``Sipp: Safe interval path planning for dynamic
  environments,'' in \emph{2011 IEEE International Conference on Robotics and
  Automation}.\hskip 1em plus 0.5em minus 0.4em\relax IEEE, 2011, pp.
  5628--5635.

\bibitem{kretz2018some}
T.~Kretz, J.~Lohmiller, and P.~Sukennik, ``Some indications on how to calibrate
  the social force model of pedestrian dynamics,'' \emph{Transportation
  research record}, vol. 2672, no.~20, pp. 228--238, 2018.

\bibitem{trautman2013robot}
P.~Trautman, J.~Ma, R.~M. Murray, and A.~Krause, ``Robot navigation in dense
  human crowds: the case for cooperation,'' in \emph{2013 IEEE international
  conference on robotics and automation}.\hskip 1em plus 0.5em minus
  0.4em\relax IEEE, 2013, pp. 2153--2160.

\bibitem{trautman2017sparse}
P.~Trautman, ``Sparse interacting gaussian processes: Efficiency and optimality
  theorems of autonomous crowd navigation,'' in \emph{2017 IEEE 56th Annual
  Conference on Decision and Control (CDC)}.\hskip 1em plus 0.5em minus
  0.4em\relax IEEE, 2017, pp. 327--334.

\bibitem{long2017deep}
P.~Long, W.~Liu, and J.~Pan, ``Deep-learned collision avoidance policy for
  distributed multiagent navigation,'' \emph{IEEE Robotics and Automation
  Letters}, vol.~2, no.~2, pp. 656--663, 2017.

\bibitem{tai2018socially}
L.~Tai, J.~Zhang, M.~Liu, and W.~Burgard, ``Socially compliant navigation
  through raw depth inputs with generative adversarial imitation learning,'' in
  \emph{2018 IEEE International Conference on Robotics and Automation
  (ICRA)}.\hskip 1em plus 0.5em minus 0.4em\relax IEEE, 2018, pp. 1111--1117.

\bibitem{tai2017virtual}
L.~Tai, G.~Paolo, and M.~Liu, ``Virtual-to-real deep reinforcement learning:
  Continuous control of mobile robots for mapless navigation,'' in \emph{2017
  IEEE/RSJ International Conference on Intelligent Robots and Systems
  (IROS)}.\hskip 1em plus 0.5em minus 0.4em\relax IEEE, 2017, pp. 31--36.

\bibitem{long2018towards}
P.~Long, T.~Fan, X.~Liao, W.~Liu, H.~Zhang, and J.~Pan, ``Towards optimally
  decentralized multi-robot collision avoidance via deep reinforcement
  learning,'' in \emph{2018 IEEE International Conference on Robotics and
  Automation (ICRA)}.\hskip 1em plus 0.5em minus 0.4em\relax IEEE, 2018, pp.
  6252--6259.

\bibitem{haarnoja2018soft}
T.~Haarnoja, A.~Zhou, P.~Abbeel, and S.~Levine, ``Soft actor-critic: Off-policy
  maximum entropy deep reinforcement learning with a stochastic actor,'' in
  \emph{International conference on machine learning}.\hskip 1em plus 0.5em
  minus 0.4em\relax PMLR, 2018, pp. 1861--1870.

\bibitem{brito2021go}
B.~Brito, M.~Everett, J.~P. How, and J.~Alonso-Mora, ``Where to go next:
  Learning a subgoal recommendation policy for navigation in dynamic
  environments,'' \emph{IEEE Robotics and Automation Letters}, vol.~6, no.~3,
  pp. 4616--4623, 2021.

\bibitem{yu2020spatio}
C.~Yu, X.~Ma, J.~Ren, H.~Zhao, and S.~Yi, ``Spatio-temporal graph transformer
  networks for pedestrian trajectory prediction,'' in \emph{European Conference
  on Computer Vision}.\hskip 1em plus 0.5em minus 0.4em\relax Springer, 2020,
  pp. 507--523.

\bibitem{kingma2014adam}
D.~P. Kingma and J.~Ba, ``Adam: A method for stochastic optimization,''
  \emph{arXiv preprint arXiv:1412.6980}, 2014.

\end{thebibliography}
\end{document}